\documentclass[12pt,a4paper]{article}
\usepackage{arxiv}
\usepackage[english]{babel}
\usepackage[T1]{fontenc}
\usepackage{amsmath,amssymb}
\usepackage{graphicx}
\usepackage{caption}
\usepackage{placeins}
\usepackage{microtype}
\usepackage{natbib}
\setcitestyle{authoryear,round,semicolon,aysep={},yysep={,}}

\usepackage{xurl}
\usepackage[hidelinks,unicode]{hyperref}
\hypersetup{pdftitle={Benchmarking Large Language Models for Linguistic Annotation},pdfsubject={Linguistic annotation and large language model benchmarking}}
\usepackage{newunicodechar}
\newunicodechar{α}{\ensuremath{\alpha}}
\newunicodechar{κ}{\ensuremath{\kappa}}
\newunicodechar{ᵉ}{\textsuperscript{e}}
\newunicodechar{ᵗ}{\textsuperscript{t}}
\newunicodechar{ʰ}{\textsuperscript{h}}

\title{Benchmarking Large Language Models for Linguistic Annotation}
\author{ \href{https://orcid.org/0000-0001-5088-3129}{\includegraphics[scale=0.06]{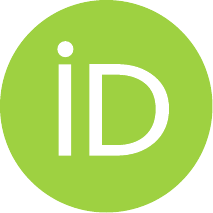}\hspace{1mm}Ondřej Tichý}
	Department of Linguistics\\
	Faculty of Arts\\
	Charles University \\
	\texttt{ondrej.tichy@ff.cuni.cz} \\
	\And
	\href{https://orcid.org/0000-0003-3392-1028}{\includegraphics[scale=0.06]{orcid.pdf}\hspace{1mm}Anna Marklová}
    \And
	\href{https://orcid.org/0000-0001-6554-0639}{\includegraphics[scale=0.06]{orcid.pdf}\hspace{1mm}Barbora Bulantová}
    \And
	{\hspace{1mm}Magdalena Titlbachová} 
    \And
    \href{https://orcid.org/0000-0001-8605-1199}{\includegraphics[scale=0.06]{orcid.pdf}\hspace{1mm}Jiří Milička}
}
\date{}

\begin{document}
\maketitle
\begin{abstract}
Assessing large language models for linguistic annotation requires task-specific evaluation under documented conditions. We present \emph{dhAIBench}, a Python benchmarking and annotation tool with a web interface for exploring and sharing results. It records prompts, configurations and predictions, enabling comparisons of annotation quality, confidence calibration, repeatability and estimated cost. We demonstrate the framework and the associated methodology through 293 runs of 48 models on nine linguistic annotation tasks spanning morphology, syntax, semantics, discourse-pragmatics, spelling normalization and learner-language correction. Models achieve high agreement with reference annotations on several tasks, including those on non-standardized, historical varieties as well as requiring contextual interpretation. The results support evaluating models against specific linguistic tasks, rather than relying on general rankings. To this end, we make the current and future benchmarks publicly available at \href{http://dhaibench.silent3.ff.cuni.cz/}{dhaibench.silent3.ff.cuni.cz}.
\end{abstract}


\FloatBarrier

\section{Introduction}\label{introduction}

A substantial portion of linguistic research depends on decisions about what a form means, which grammatical category it instantiates and how it functions in context -- in other words, on linguistic annotation. Making these decisions at corpus scale requires an automated annotation procedure whose outputs can be inspected and whose limitations are understood. Automated annotation has traditionally relied on dedicated natural language processing (NLP) tools, which can offer computational efficiency and reproducible results for well-defined tasks. LLMs offer a practical means of applying instructions expressed in ordinary language, including instructions for categories that are less strictly defined or that apply to non-standardized varieties and therefore poorly served by existing software, without requiring a separately trained model for each task. Their usefulness therefore raises an empirical question: under which conditions do their annotations provide dependable evidence for a linguistic analysis?

Studies of genre identification, semantic classification and multilingual tasks illustrate the potential of LLM annotation and the variation in performance between applications {\citep{Kuzman2023-es,Lai2023-uv,Koeva2024-sk}.} Assessing suitability for a new project nevertheless remains difficult. Reports of LLM use in linguistic research do not always specify model versions, prompts, generation settings and evaluation procedures in sufficient detail for the results to be easily reproduced or compared. Popular benchmarking sites and general-purpose leaderboards, meanwhile, do not focus on linguistic annotation. Researchers consequently need a practical way to establish whether the annotations required for their study can be usefully and efficiently automated using LLMs, choose an appropriate model, and carry out the annotation without necessarily having advanced programming skills.

Our resources do not permit a comprehensive or representative survey of linguistic annotation tasks across all available LLMs. We have therefore developed \emph{dhAIBench} around three complementary resources: a benchmarking tool that researchers can use with their own data; a web application for exploring and analyzing the benchmarks produced by the tool; and a public website that collects and presents shared results. The tool supports both benchmarking against reference annotations and using the selected model to annotate additional research data. Together, these resources aim to make task-specific evaluation and subsequent annotation more accessible, while preserving the information needed to interpret and compare the results.

To establish an initial collection and demonstrate the workflow, we contribute benchmarks for nine tasks covering several levels of linguistic description, from morphology and syntax to semantics and discourse-pragmatics, alongside historical text normalization and learner-language correction. The tasks concern English to make the examples accessible to a broad readership, while giving attention also to less commonly evaluated varieties. This selection illustrates uses of the framework within the resources available to us.

These benchmarks cannot be used to establish a general leaderboard of models for linguistic annotations and they are unlikely to match the precise data, categories and analytical requirements of most other projects. They can nevertheless provide provisional points of comparison. A researcher considering a morphological annotation task, for example, can examine the accuracy achieved on another morphological task, the models used and any reported time and monetary costs. Such comparisons can inform an initial choice of models and a judgement about feasibility; the acceptable level of accuracy must still be determined for the intended analysis. We therefore encourage researchers to benchmark their own tasks and to share their results for inclusion on the website.

Our longer-term aim is to broaden the collection through contributions from other researchers. Even an expanded collection will remain incomplete: models are continually introduced and retired. Retirement can prevent a new task from being evaluated with an older model, while testing every existing task with each newly available model would exceed the resources of the project. The collection is thus intended to grow as a shared record of task-specific evaluations. This chapter describes the tools and initial benchmarks, and discusses how their results can support decisions about LLM annotation.

\FloatBarrier

\section{Linguistic Annotation and LLM benchmarking}\label{linguistic-annotation-and-llm-benchmarking}

Linguistic annotations are explicit decisions about particular instances of language use that depend on the annotation unit, category definitions, available context and conventions for ambiguous cases. Agreement with reference annotations therefore measures conformity to an operational definition; it does not independently establish the validity of the underlying analysis. Research on inter-annotator agreement distinguishes these questions {\citep{Artstein2008-ek},} while work on human label variation shows that disagreements can reflect defensible differences in interpretation rather than annotator error alone {\citep{Plank2022-rv}.} These considerations apply equally to human annotation, conventional NLP tools and LLMs.

\subsection{From dedicated NLP tools to prompted annotation}\label{from-dedicated-nlp-tools-to-prompted-annotation}

Established NLP pipelines commonly implement annotation through rules, lexical resources or models trained for particular tasks. Where suitable tools exist, they can provide efficient processing, consistent output structures and reproducible application of a specified annotation scheme. Their suitability nevertheless depends on the relationship between their original design and the researcher's data. Applying an existing tagger to historical correspondence, for example, requires consideration of differences in spelling, grammatical usage and annotation conventions. Adapting a pipeline to new categories or varieties may require programming, additional lexical resources or labelled training data.

LLMs belong within NLP, but their use through prompting changes how researchers can specify an annotation task. Category definitions, decision rules, examples and contextual information can be supplied in ordinary language without training a separate model for each application. Brown et al. {\citeyearpar{Brown2020-wb}} demonstrated this capacity to specify tasks and examples through textual interaction without updating model parameters. For linguistic research, this makes it possible to experiment with categories that are absent from existing annotation tools. It does not remove the need to define those categories carefully or to obtain reference annotations for evaluation.

Prompting also makes the instructions part of the evaluated system. Changing a category description, adding examples or supplying grammatical metadata changes the information available to the model. Even changes that preserve the intended meaning can affect performance: Sclar et al. {\citeyearpar{Sclar2024-ei}} document substantial sensitivity to prompt formatting in few-shot classification. A benchmark consequently evaluates a model under a particular prompting procedure.

The relationship between linguistic conventions and computational implementation is particularly apparent in historical text processing. Bollmann {\citeyearpar{Bollmann2019-eq}} compares normalization systems based on several methodological traditions and highlights difficulties arising from different datasets and evaluation procedures. Hämäläinen et al. {\citeyearpar{Hamalainen2019-hq}} show that lexical resources can improve neural normalization through filtering candidate outputs. Such findings also motivate treating dictionaries, normalization rules and validators as components of an annotation system whose contribution must be recorded alongside that of the model.

\subsection{Benchmark design and reproducibility}\label{benchmark-design-and-reproducibility}

A benchmark for linguistic annotation should establish how well a specified annotation procedure meets the requirements of an intended research task. Its design must define the linguistic phenomenon and annotation categories, the source and sampling of material, and the reference annotations against which predictions are assessed. The evaluated procedure includes both the model and the conditions of its use: the instructions, contextual information and examples supplied to it, together with any constraints or subsequent validation of its outputs. Benchmark scores must therefore be interpreted in relation to these conditions.

Evaluation should also reflect the practical requirements of linguistic research. Agreement with reference annotations is central, but aggregate accuracy alone may conceal poor performance on categories important to the analysis. Coverage, consistency between runs, the reliability of confidence estimates and computational costs provide additional evidence about whether an annotation procedure is suitable for use {\citep{Liang2022-lp}.}

\FloatBarrier

\section{Methodology}\label{methodology}

We use the \emph{node} as the annotation unit: a predefined word, multiword expression or skip-gram, supplied with context and optional metadata in a format resembling a keyword-in-context (KWIC) corpus export. Each request concerns one node, without conversational history from other items. This simplifies benchmarking: each prediction can be compared directly with the reference label for the same node. Annotating complete texts may additionally require matching predicted and reference spans, whose boundaries can differ. Our procedure separates annotation from instance detection and segmentation, so scores measure agreement on supplied targets rather than the ability to identify instances in unprocessed text.\footnote{Detection or segmentation can also be evaluated over predefined units using labels to encode target membership or segment boundaries. Users can, for example, process a text word by word but always with the whole text supplied for context and with annotations representing the desired segmentation.} Repeating instructions and context, however, increases resource use.

The technical description below explains how \emph{dhAIBench} records inputs, settings and outputs and supports evaluation and review of individual predictions.

\subsection{Technical description}\label{technical-description}

The implementation has three main components: a Python benchmarking script with a browser-based configuration interface, optional external validators and a static metrics dashboard -- all freely available from a GitHub repository.\footnote{The public repository is at \url{https://github.com/otichy/dhAIBench}, the public instance of the web app with our benchmarking results is at \url{https://dhaibench.silent3.ff.cuni.cz/}}

\subsubsection{Benchmarking script}\label{benchmarking-script}

The script reads semicolon-delimited CSV files with at least ID and node fields. Optional leftContext and rightContext fields supply the immediate surroundings, while info supplies task-relevant metadata. The rest of fields are never sent to the model. An optional truth column, or a separate ID--label file, supplies the reference labels for evaluation. This separation permits scoring against annotations that the model does not see for the target item.\footnote{By default, the runs are zero-shot (no example of a correct annotation is provided to the model), but the user can either include examples in the prompt or can set a number of input truth labels to be provided to the model.}

Other input columns are retained in the output (e.g. to preserve a complete multi-column corpus export, user comments etc.), but do not enter the process in any way.

For the standard, single-item workflow, the script constructs a task-level system prompt and an item-level user prompt. The system prompt is based on user task-specific annotation instructions and general instructions on the prompt-response exchange, while the user prompt is generated from the input data per each annotation item.\footnote{The division between system and user prompt allows for caching the system prompt, which remains the same for all items of a task, and leads to more efficient and cheaper runs.} The node is marked explicitly, and a classification\_target field identifies either the complete node or marked internal spans (e.g. in cases of discontinuous linguistic structures). The requested JSON response from the model contains the prediction, an optional explanation and confidence (as estimated by the model in its answer and/or as token logprobs), and fields echoing the target and its source. Checking the echoed span helps detect a well-formed answer about the wrong piece of text; it does not establish that the linguistic label is correct.\footnote{User can set a number of automatic retries in cases the answer is malformed or the node is not echoed correctly.}

In this workflow, each target is classified without conversational memory of previously classified items.

The configuration records model and provider identifiers, generation controls and operational settings such as concurrency, request spacing, caching, timeouts and retries. Supported controls include temperature, top-p and, where available, top-k, reasoning effort and provider-specific thinking effort settings. User can also set descriptive metadata like task name, description and tags that are useful for later analysis.

The complete details of the script's parametrization is part of its documentation. Because the settings are relatively complex, there is also a web-based configuration where the user can set the run up using a user-friendly GUI and then copy the resulting command to run the script.

Predictions are saved in a CSV containing the input and response fields. Newline-delimited JSON (NDJSON) prompt logs record per-attempt information, while metrics and all the run parameters are saved in JSON and can be recomputed without issuing new model requests. Resume mode reuses saved rows by ID to support recovery from interruption but also allows to re-prompt any records so labeled as unclassified.\footnote{Records can be labeled as unclassified by the model due to technical issues (malformed answer, incorrect node identification), but also because the model was not able to decide what label to apply. Since the models behavior is usually stochastic, this may be alleviated by a re-prompt.}

\subsubsection{Validators}\label{validators}

The script can start an external validator and communicate with it through one JSON message per attempt over standard input and output. A validator is a deterministic process (e.g. a Python script) that can accept a prediction, optionally normalize it, request a retry with a restricted candidate label list and explanatory message, or abort the run. Retries reconstruct the base prompt and append the validator feedback.

This architecture is useful e.g. for tasks with large, allowed label sets like lemmatization or normalization where including a complete dictionary in every prompt would be costly or impractical. Candidate generation can use e.g. lexical resources, string distance or part-of-speech information. It combines linguistic constraints with model selection while keeping deterministic transformations outside the language model. However, an accepted dictionary entry can still be the wrong lemma in context. Validator acceptance and linguistic correctness are separate outcomes.

\subsubsection{Metrics web app}\label{metrics-web-app}

The browser dashboard is designed to be used locally by individual researchers to analyze their runs or online to share the interactive results publicly. It can load a local metrics folder or a server collection and filter runs by task, model, tags and time. Its views expose accuracy and other label metrics\footnote{Metrics definitions:

  \textbf{Accuracy}: exact-match rate, i.e. the share of examples where predicted label = gold label.

  \textbf{Brier score}: mean squared error between confidence and correctness; lower is better.

  \textbf{Expected Calibration Error (ECE)}: average gap between stated confidence and actual accuracy across confidence bins; lower is better.

  \textbf{Maximum Calibration Error (MCE)}: worst single confidence-bin gap; lower is better.

  \textbf{Clustering distance}: raw disagreement rate between runs, used for similarity trees rather than leaderboard ranking.

  \textbf{Cohen's kappa}: accuracy corrected for chance agreement given the label distribution;~1~= perfect agreement,~0~= chance-level,~\textless0~= worse than chance.

  \textbf{Confusion matrix}: where the model systematically confuses one label for another.

  \textbf{Cross-model α}: how similarly different models classify the same task variant.

  \textbf{Macro precision / recall / F1}: class-balanced averages of the per-label scores, so rare labels count equally with frequent ones.

  \textbf{Repeat α}: stability of the same model across repeated runs on the same task variant.}, calibration, token usage, estimated prices and repeated-run or cross-model agreement in a number of visualizations: grouped chart, scatterplot, sortable table, radar graph, agreement tree and a customizable leaderboard. Links from individual run's details connect summary values to metric files and, where published, prediction files and logs.

Grouped summaries require explicit interpretation. In the default model grouping, repeated runs are averaged within a task and the task means are then averaged. Models evaluated on different task sets therefore have different denominators. The customizable leaderboard can require wha common set of tasks and explicit weights, which makes its composite score more interpretable, but comparable task names alone do not automatically establish identical prompts, validators or dataset versions, the user can control for this using tags.

\begin{figure}[p]
\centering
\includegraphics[width=.98\linewidth]{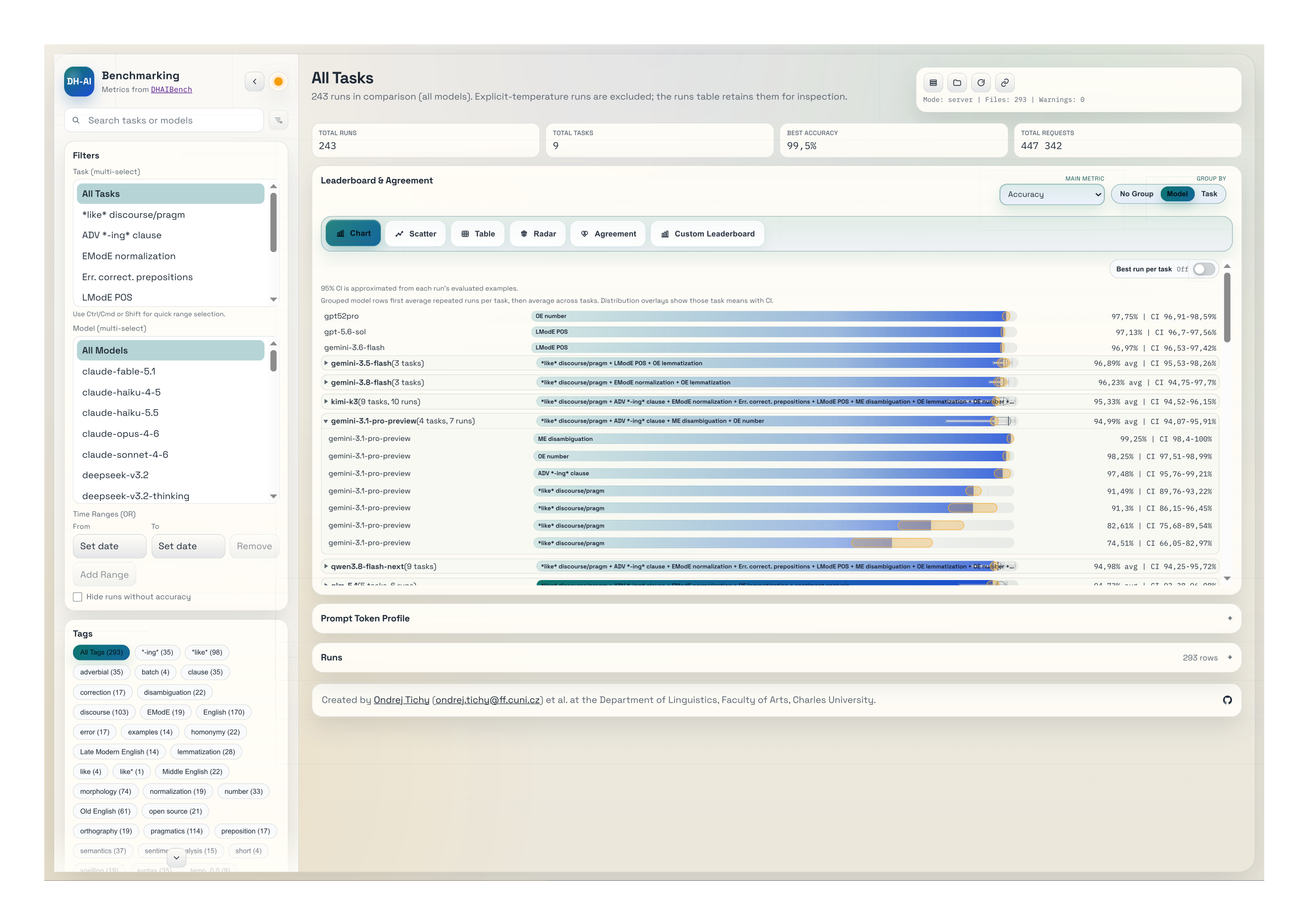}
\caption{Metrics web app main dashboard}
\label{fig:01}
\vspace{1em}
\begin{minipage}[t]{.59\linewidth}
\vspace{0pt}\centering
\includegraphics[width=\linewidth]{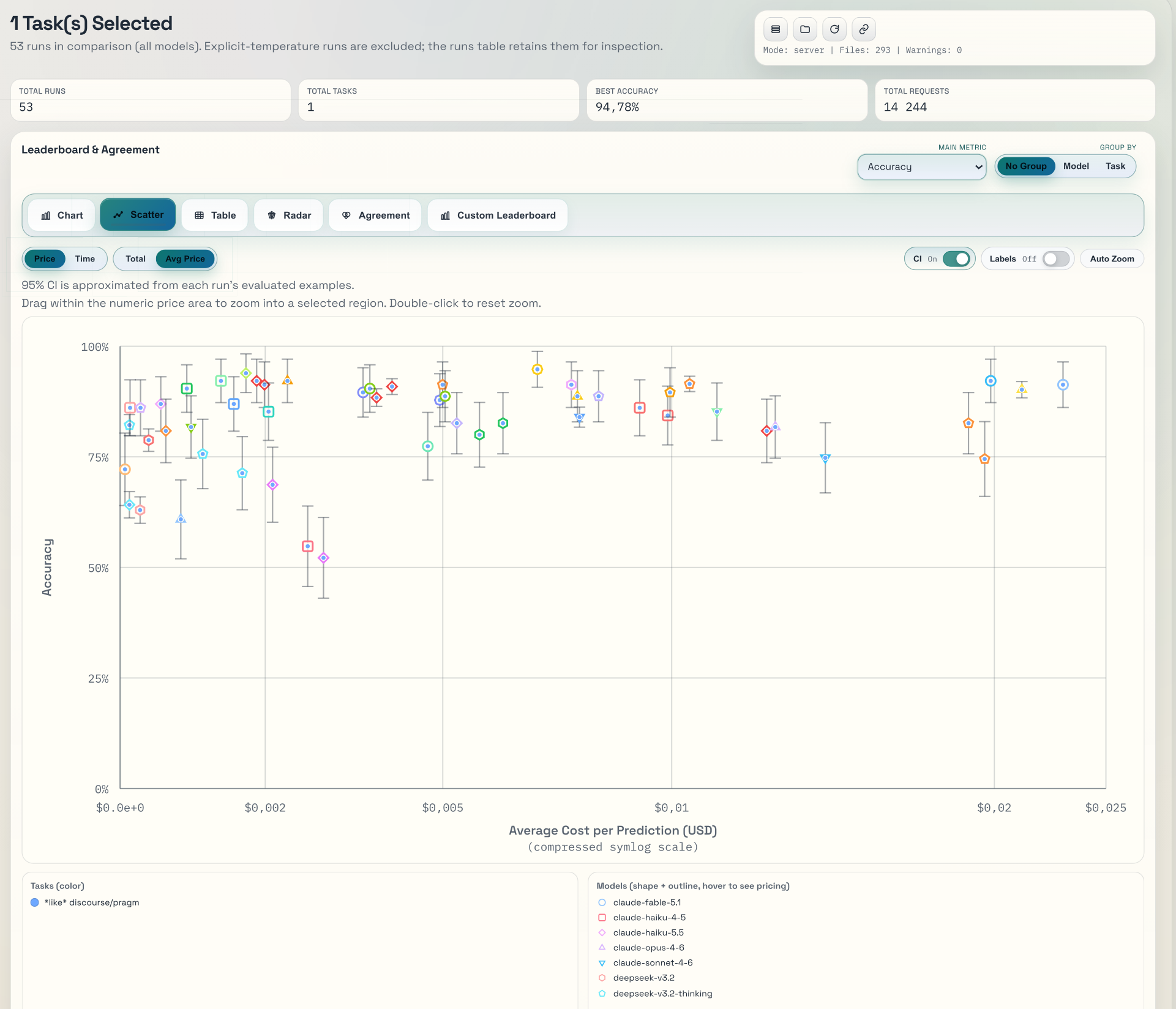}
\caption{Metrics web app scatterplot showing Average cost per prediction vs. Accuracy for one task}
\label{fig:02}
\end{minipage}\hfill
\begin{minipage}[t]{.37\linewidth}
\vspace{0pt}\centering
\includegraphics[width=\linewidth]{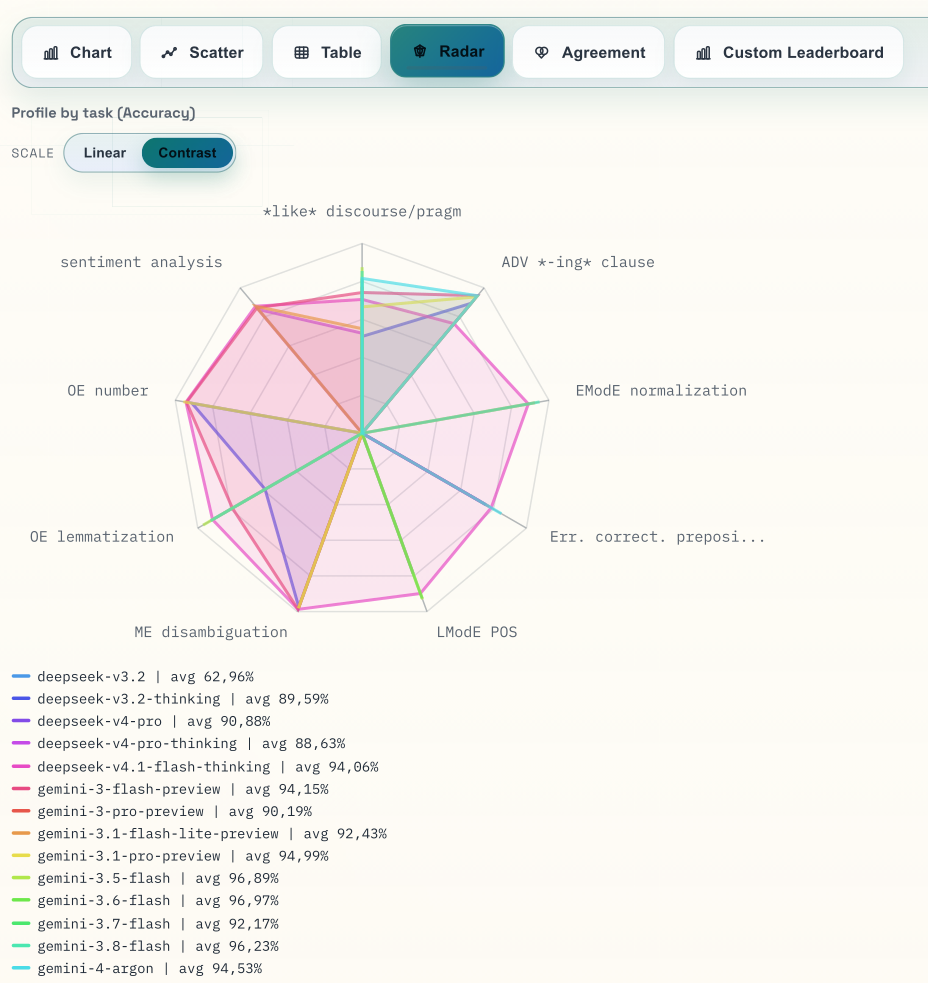}
\caption{Metrics web app radar chart model comparison}
\label{fig:03}
\end{minipage}
\end{figure}

Agreement is calculated from predictions aligned by ID, with task fingerprints incorporating input fields and reference labels. The implementation distinguishes repeated runs of the same provider--model combination from agreement between different models. Cross-model summaries can use the latest run or the run with the highest accuracy.\footnote{Note that this can overstate expected performance for frequent re-runs.}

\begin{figure}[htbp]
\centering
\includegraphics[width=.88\linewidth,height=.78\textheight,keepaspectratio]{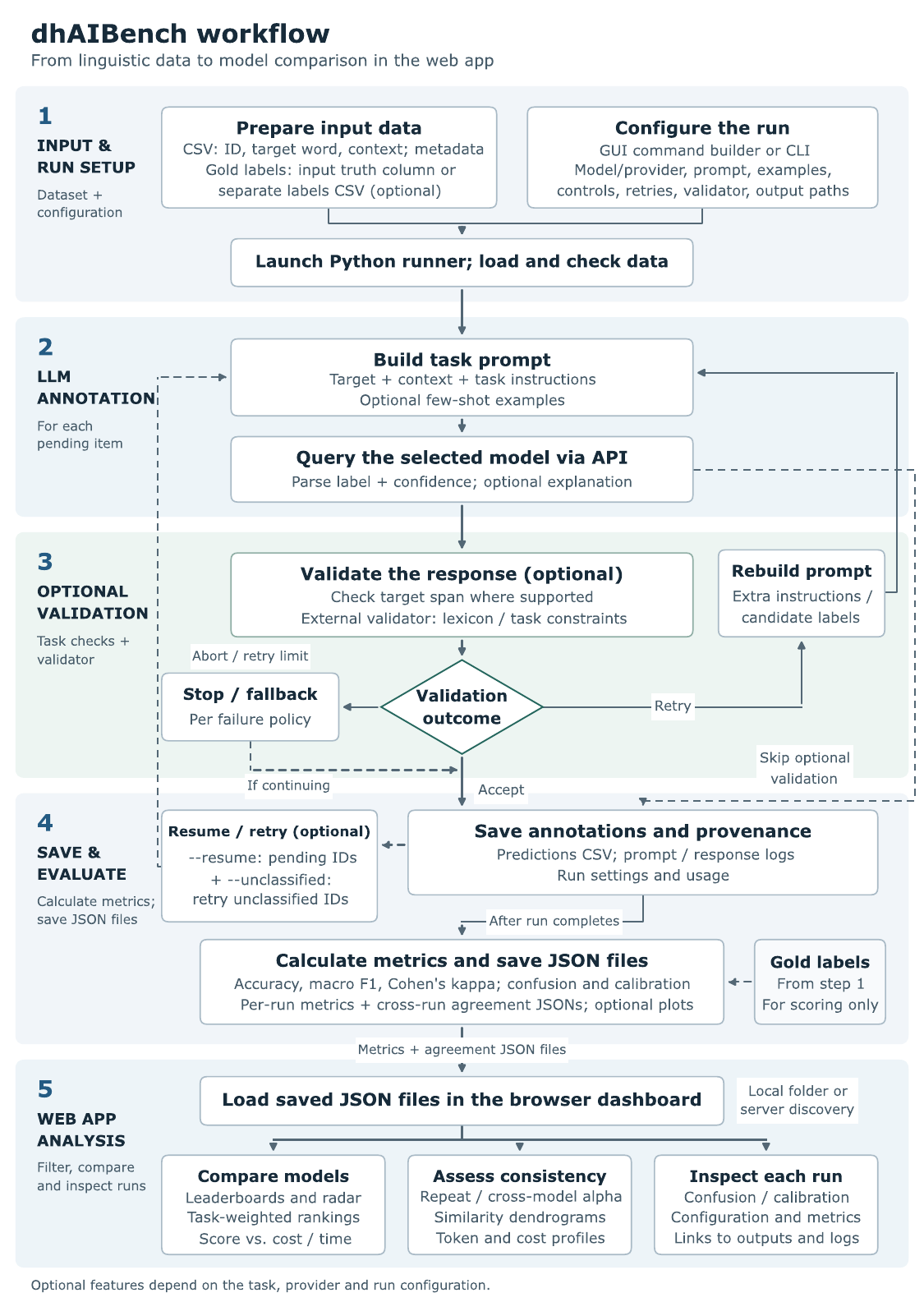}
\caption{Workflow flowchart}
\label{fig:04}
\end{figure}

\FloatBarrier

\section{Datasets}\label{datasets}

The datasets were selected pragmatically on the basis of three criteria: 1. They involved linguistic annotation tasks that were not reliably addressed by conventional NLP tools. 2. The data and their reference annotations were available to the authors, often through their own research. 3. The authors had sufficient knowledge of the annotation procedures to formulate prompts similar to instructions for human annotators and to analyze the annotations produced by \emph{dhAIBench}.

The prompts were formulated in natural language and ranged from very short and simple (for the Sentiment analysis task\footnote{Prompt: ``You will analyze an excerpt representing an online movie review and decide whether the excerpt represents a positive or a negative sentiment of the reviewer towards the movie. In case of a positive sentiment, return `positive', in case of a negative sentiment, return `negative'.''}) to long and complex -- the LModE PoS tagging task prompt included the complete CLAWS tagset and a large number of examples while the Discourse-pragmatice uses of \emph{like} task prompt comprised of instructions for a hierarchical multi-step annotation process.

\subsection{\texorpdfstring{T1: Discourse-pragmatic uses of \emph{like}}{T1: Discourse-pragmatic uses of like}}\label{t1-discourse-pragmatic-uses-of-like}

The dataset contains 115 contextualized instances of \emph{like}, corresponding to the inter-rater validation sample in Raušová's {\citeyearpar{Rausova2026-ke}} \emph{Discourse-pragmatic like: An empirically grounded functional taxonomy}. The examples originate from two corpora of informal spoken British English: the demographically sampled component of the BNC1994 and the Spoken BNC2014. The sample was selected to cover the taxonomy's functional distinctions and was annotated by two independent raters using a decision tree. Each benchmark item presents a target occurrence alongside its surrounding dialogue. Classification draws primarily on omissibility, syntactic position, and syntactic boundedness - that is, whether \emph{like} is structurally associated with the material preceding or following it.

The benchmark distinguishes four discourse-pragmatic functions. Discourse markers (40 instances) are syntactically unbound and organize discourse or manage disfluency, as in ``Board games today are insane like I\textquotesingle ve my friends bought me {[}...{]}''. Clause-medial pragmatic markers (25 instances) scope forward over following material, highlighting information or signaling that its interpretation requires enrichment or loosening: ``my friends have got \emph{like} nose rings and stuff''. Clause-final pragmatic markers (25 instances) scope backward, conveying emphasis, qualifying preceding material, or inviting confirmation, as in ``I'm as stubborn, stubborn as a mule \emph{like}.'' Quotative markers (20 instances) introduce reported speech, thoughts, feelings, or attitudes, commonly within \emph{be like}: ``the guy was \emph{like} no no no you paid this village \ldots''. These examples are drawn from the dataset, with target occurrences italicized.

A fifth label, unclassified (5 instances), covers propositional uses outside the discourse-pragmatic taxonomy, such as the verb in ``I don't really \emph{like} skype.'' Although the paper's taxonomy includes three hierarchical levels, the benchmark retains only the four main discourse-pragmatic categories and this propositional contrast class. The task consequently evaluates broad functional classification of \emph{like} in conversational context, rather than the finer distinctions among its pragmatic subfunctions.

\subsection{\texorpdfstring{T2: Adverbial \emph{-ing} clauses}{T2: Adverbial -ing clauses}}\label{t2-adverbial--ing-clauses}

The ADV-ing dataset contains 318 manually identified English adverbial \emph{-ing} clauses adapted from Lávička's {\citeyearpar{Lavicka2024-op}} thesis \emph{Adverbial -ing Clauses in Newswriting: A Diachronic Comparison}. The source material comprises 108 \emph{Guardian} news reports on politics and current affairs, drawn from printed newspapers published in 1953 and online articles published in 2023. The dataset includes 126 instances from the earlier period and 192 from the later period.

The annotation scheme distinguishes three construction types. Free adjuncts (FA; 216 instances) lack an overt subject, typically sharing their understood subject with the matrix clause, as in ``\emph{Speaking to ABC News}, the governor said search-and-rescue efforts were ongoing.'' and may also be introduced by an optional conjunction such as \emph{while}. Adverbial verbal gerunds (AVG; 81 instances) are governed by an obligatory preposition, as in ``A bee-keeper restored order \emph{by removing the jacket, complete with swarm, in a sack}.'' Absolute constructions (AC; 21 instances) have their own overt subject and may be introduced by \emph{with}, as in ``Areas along the Emilia-Romagna coastlines were also affected by the floods, \emph{with debris washing up on beaches}.'' These examples are drawn from the dataset, with the target constructions italicized.

\subsection{T3: Normalization of Early Modern English printed text}\label{t3-normalization-of-early-modern-english-printed-text}

The dataset contains 7,489 token-level instances adapted from Titlbachová's {\citeyearpar{Titlbachova2025-fr}} thesis, \emph{Normalization of non-standardized varieties of Early Modern English using AI language models}. Its 15 text samples follow a sampling framework combining three periods: 1500--1550, 1550--1600, and 1600--1650, with five genres: literary, religious, administrative, and educational texts, and personal correspondence. One text represents each period--genre combination, with excerpts of approximately 500 words. Printed texts were selected through the \emph{Universal Short Title Catalogue} and obtained from the \emph{Early Print Library}; correspondence comes from the \emph{Parsed Corpus of Early English Correspondence}. Each benchmark item supplies an original token, surrounding context, source metadata (title, author publication date), and a manually standardized reference form.

The annotation task is spelling normalization, it regularizes historical orthography as in \emph{fallynge \textgreater{} falling}, expands abbreviations such as \emph{yᵉ \textgreater{} the}, but preserves word order and morphology (e.g. \emph{hath} is retained), prohibits adding, deleting or replacing words (e.g. obsolete vocabulary with modern synonyms) and retains punctuation and capitalization.

The dataset includes both forms requiring normalization and forms retained unchanged: 1,941 source--reference pairs differ, while 5,548 are identical. Successful normalization consequently requires deciding when intervention is appropriate as well as selecting the correct spelling.

\subsection{T4: Error-correction of prepositions in learners' language}\label{t4-error-correction-of-prepositions-in-learners-language}

The dataset was compiled by Todorova {\citeyearpar{Todorova2023-vl}} to investigate the use of the prepositions \emph{in}, \emph{on} and \emph{at} by advanced learners of English with Spanish as their L1. The data was drawn from the Spanish subcorpus of the Louvain International Database of Spoken English Interlanguage {\citep{Gilquin2010-gv},} which comprises 50 orthographically transcribed interviews consisting of a monologue, an informal conversation and a picture-description task.

The dataset contains 1,284 instances of the target prepositions presented in their immediate context. Each instance is annotated for accuracy, and a correction is provided where the preposition use was judged to be erroneous.

\subsection{T5: Part-of-speech tagging of Late Modern English correspondence}\label{t5-part-of-speech-tagging-of-late-modern-english-correspondence}

The LModE POS dataset contains 5,681 annotated tokens drawn from the 15-letter test sample described by Saario-Ramsay et al. in \emph{My grat destress -- POS tagging 18th-century English letters} {\citep{Saario-Ramsay_undated-jh}.} The letters originate from the \emph{Corpus of Early English Correspondence Extension} (CEECE) and span 1690--1784, representing male and female writers across social ranks, from paupers to the nobility. The material retains historical spelling and capitalization, including forms such as \emph{grat} (`great') and \emph{destress} (`distress').

The annotation follows the fine-grained CLAWS7 scheme, with 139 distinct reference labels represented in the supplied dataset, including punctuation and multiword-expression tags. The scheme distinguishes grammatical categories together with morphological and functional properties. For example, \emph{letter} receives NN1 (singular common noun), whereas \emph{thoughts} receives NN2 (plural common noun); \emph{I} is PPIS1 (first-person singular subject pronoun), whereas \emph{me} is PPIO1 (the corresponding object pronoun). Verbal distinctions include VVD for lexical past-tense forms such as \emph{gave}, VVN for past participles such as \emph{stolen}, and VVI for infinitives such as \emph{tell}. Separate tag families distinguish forms of \emph{be}, \emph{have}, and \emph{do} from other verbs.

Multiword expressions receive ditto tags, which combine a grammatical category with numeric suffixes identifying the expression's length and each component's position. Thus, \emph{at least} is annotated at\_RR21 least\_RR22, treating the two words jointly as an adverb, while \emph{so that} receives so\_CS21 that\_CS22, marking a subordinating conjunction. This convention also accommodates historically separate spellings, such as after\_RT21 words\_RT22 (`afterwards'). Punctuation marks are included as classification targets and generally receive the mark itself as their label. The underlying gold standard covers both tokenization and tagging: annotations were prepared by one annotator, checked by a second, and unclear cases resolved jointly.

\subsection{T6: Semantic disambiguation of Middle English nouns}\label{t6-semantic-disambiguation-of-middle-english-nouns}

The dataset originates in Tichý and Přečková's {\citeyearpar{TichyUnknown-ks}} research on Morphological countability in Middle English. It is an export of all potential spelling variants of four nouns: \emph{loaf, man, meat} and \emph{son} from \emph{The Penn-Helsinki Parsed Corpus of Middle English}, second edition (PPCME2). Since Middle English spelling is highly varied and the corpus is not lemmatized, finding specific lexical items is difficult, because tokens can be formally highly ambiguous. For example, lemma \emph{loaf} (\emph{lof}) can appear in forms like \emph{loues} or \emph{lof} which are in turn potential spelling of lemmas \emph{love} and \emph{leaf} respectively.

The models were tasked with disambiguating the instances by annotating all actual occurrences of the four target words with the corresponding lemma or returning \emph{other}, in case the instance represent a different lexical item.

The input data also supplied the model with text metadata like name of the text, Middle English dialect, genre and dating to inform the model's decision.

\subsection{T7: Lemmatization of Old English}\label{t7-lemmatization-of-old-english}

This dataset was benchmarked in partnership with Faulkner \& Magnanti {\citeyearpar{Faulkner2027-wj}} in view of producing a new lemmatized Old English corpus. It contains 2,846 tokens of OE and Latin\footnote{Latin tokens are ignored and labeled as foreign\_word.} from \emph{Ælfric's Catholic Homilies} exported with their immediate context\footnote{50 tokens left \& right.} from \emph{The Lemmatized York-Helsinki Parsed Corpus of Old English Prose}, Beta version 12 {\citep{Taylor2025-xv}.}

The task is contextual lemmatization: assigning the appropriate OE lexical headword to a form encountered in running text. It groups inflectional and orthographic variants under shared lemmas e.g. \emph{com \textgreater{} cuman}, \emph{bocum \textgreater{} boc}, \emph{þisum \textgreater{} þes} and regularizes orthographic variation, as in \emph{syþþan \textgreater{} siþþan}. Context determines the appropriate analysis of ambiguous forms: for example, \emph{þa} receives either \emph{se} when functioning as a demonstrative or \emph{þa} in temporal uses. The PoS and some morphological information present in the original corpus are also supplied to the model.

This annotation task makes use of a validator (see section \ref{validators}), which checks each predicted lemma against a reference lexicon derived from the \emph{Bosworth-Toller Anglo-Saxon Dictionary Online} {\citep{Tichy2024-lz}.} Predictions are matched against the dictionary headwords filtered by the corresponding PoS information, exact matches including those with pre-defined orthographic equivalents (e.g., upper/lowercase, þ/ð) are accepted and the validator stops. Otherwise, near-match candidate lemmas are ranked by a weighted edit distance that accommodates OE spelling variation, including vowel alternations, doubled letters, final \emph{-e}, and the \emph{ge-} prefix. The validator requests a revised prediction using the closest candidates (by default, up to 30 within a distance of 1.25). This procedure helps to enforce lexical compatibility and to lower hallucination but does not establish contextual correctness (only dictionary lemmas are allowed into the output, but they may not necessarily be the correct lemmas).

\subsection{T8: Number of Old English nouns}\label{t8-number-of-old-english-nouns}

The Old English nouns show a large degree of syncretism in their endings (not to mention their spelling variation) which makes it difficult to recognize their morphological categories based on the form of the ending alone. This dataset represents an export from \emph{The York-Toronto-Helsinki Parsed Corpus of Old English Prose} (YCOE) of forms of twelve nominal lemmas\footnote{\emph{ale, book, brother, death, flesh, hand, might, milk, name, sin, water} and \emph{wood.}} that are, due to the syncretism, ambiguous in their number. For example, \emph{handa} can be either genitive/dative sg. or nominative/accusative/genitive plural.

For Tichý's and Ripl's {\citeyearpar{TichyUnknown-my}} research on Countability in Old English, we have used the models no annotate instances of these nouns as either singular or plural and in addition to the forms and their immediate context, we have provided the models with the information on case, as annotated in the corpus.

\subsection{T9: Sentiment analysis of the Movie Review Data}\label{t9-sentiment-analysis-of-the-movie-review-data}

The sentence polarity dataset (commonly the Pang \& Lee sentence polarity dataset) is a binary sentiment-classification corpus of English movie-review snippets drawn from Rotten Tomatoes. It contains 10,662 snippets in total, evenly balanced into 5,331 positive and 5,331 negative examples, provided as two plain-text files (rt-polarity.pos and rt-polarity.neg) with one snippet per line. Each snippet is roughly a single sentence and all text is lower-cased. Labels were assigned automatically from the review's Rotten Tomatoes rating: snippets from reviews marked ``fresh'' are treated as positive and those marked ``rotten'' as negative. The dataset was introduced by Pang and Lee {\citeyearpar{Pang2005-ab}} and has become a standard benchmark for

\section{Results \& Discussion}\label{results-and-discussion}

\subsection{General results}\label{general-results}

The benchmarking conducted between November 2025 to 2026 comprises of 293 runs covering nine annotation tasks,\footnote{Coverage was uneven: 198 of the 432 possible model-task combinations were represented, and only eight models covered all nine tasks.} 406,783 evaluated annotation instances (including repeated evaluations) using 4.18 billion input and output tokens from 48 models.\footnote{claude-fable-5.1, claude-haiku-4-5, claude-haiku-5.5, claude-opus-4-6, claude-sonnet-4-6, deepseek-v3.2, deepseek-v3.2-thinking, deepseek-v4-pro, deepseek-v4-pro-thinking, deepseek-v4.1-flash-thinking, gemini-3-flash-preview, gemini-3-pro-preview, gemini-3.1-flash-lite, gemini-3.1-flash-lite-preview, gemini-3.1-pro-preview, gemini-3.5-flash, gemini-3.6-flash, gemini-3.7-flash, gemini-3.8-flash, gemini-4-argon, gemma-4-26b-a4b-it, glm-4.7, glm-5, glm-5.1, glm-5.2, glm-5.3, gpt-5.4, gpt-5.4-mini, gpt-5.4-pro, gpt-5.5, gpt-5.6-luna, gpt-5.6-sol, gpt-6-astra, gpt-6-luna, gpt-oss-120b, gpt5, gpt51, gpt52pro, gpt5mini, jev-1.13.0, kimi-k2.5, kimi-k2.6, kimi-k2.7, kimi-k3, mercury-2, qwen3.5, qwen3.6, qwen3.8-27b, qwen3.8-flash-next, qwen3.8-max.} All models except two are broadly speaking autoregressive transformers, Mercury-2 is a diffusion model and Jev-1.13.0 is reportedly a transformer-based decision model. The e-INFRA CZ research infrastructure was used as the inference provider for all the open-weights models, OpenAI and Google were used for their own models, and the rest of the models were benchmarked through OpenRouter and Requesty services. All model parameters influencing the quality of the predictions were left at their providers' defaults, except when specifically noted. All results including the parameters of each run, the prompts and the metrics\footnote{Specifically, all run parameters including the task specific part of the prompt are publicly available in the web app, the item specific part of the prompt can be reconstructed from the dhAIBench repository and the input data. However, the input, output and logs are not publicly available due to licensing of some of the datasets and to avoid benchmark contamination.} can be viewed in our online instance of the web app.

\begin{figure}[htbp]
\centering
\includegraphics[width=1.0\linewidth,height=.78\textheight,keepaspectratio]{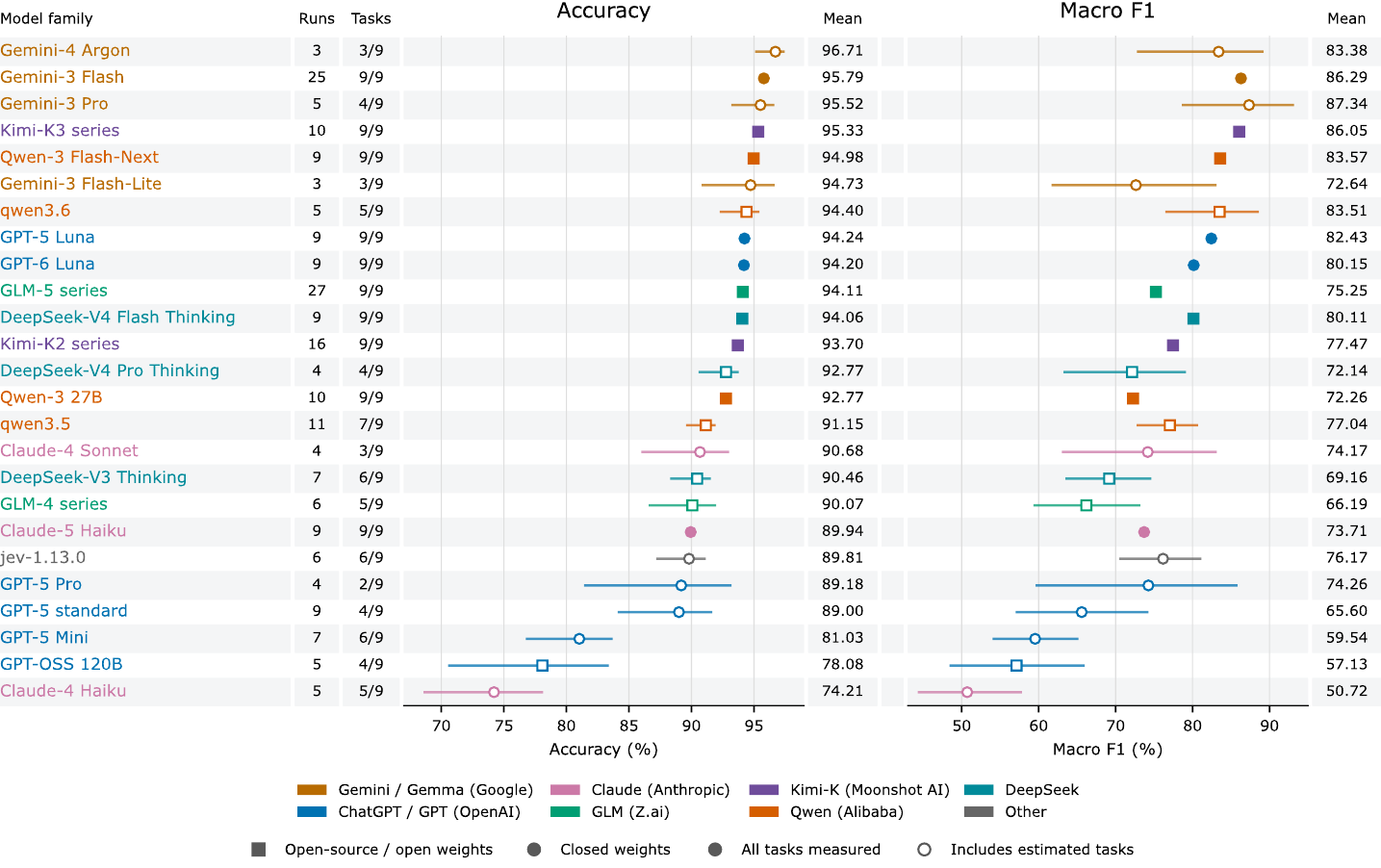}
\caption{{}Task-adjusted comparison of model families by Accuracy and Macro F1{}}
\label{fig:05}
\end{figure}

The overall model comparison is shown in Figure~\ref{fig:05}. The model subversions are grouped into families\footnote{Only families with at least three runs per task are included. Gemini-3 Flash family, for example, includes gemini-3-flash-preview, gemini-3.5-flash, gemini-3.6-flash, gemini-3.7-flash, gemini-3.8-flash.} and are made comparable (despite the uneven coverage -- not all models were benchmarked on all tasks) by calculating task-adjusted scores using a two-way regression with family and task effects, fitted on the logit scale for accuracy and the original scale for Macro F1. Regression predictions fill only missing family--task scores; observed scores remain unchanged. Whiskers show 95\% model-based prediction ranges reflecting uncertainty in the missing-task estimates.

\begin{figure}[htbp]
\centering
\includegraphics[width=1.0\linewidth,height=.78\textheight,keepaspectratio]{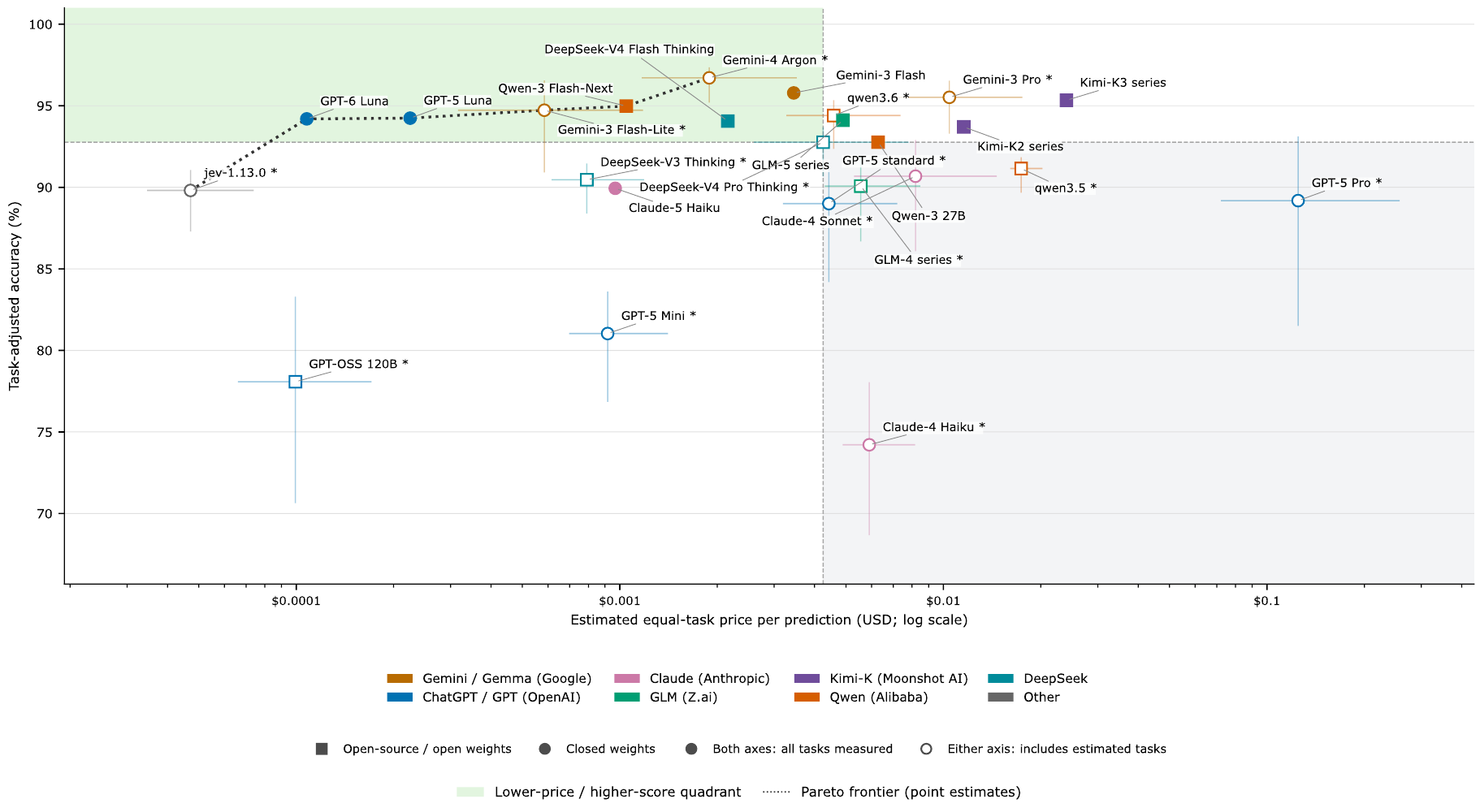}
\caption{{}Price - performance comparison.}
\label{fig:07}
\end{figure}

Our benchmarks are clearly lead by the Gemini model family, but the large open-source models like Kimi-k, Qwen, GLM and DeepSeek are not far behind. While GPT and Claude are perhaps surprisingly not the top models families here, note that the largest models of those families, Astra and Fable, are not listed here. Due to their high price, we have tested them only on T1 (see Figure~\ref{fig:07}), where claude-fable-5.1 narrowly beat gpt-6-astra as well as all the open-source models, but remained behind a much cheaper gemini-3.5-flash.

\begin{figure}[htbp]
\centering
\includegraphics[width=1.0\linewidth,height=.78\textheight,keepaspectratio]{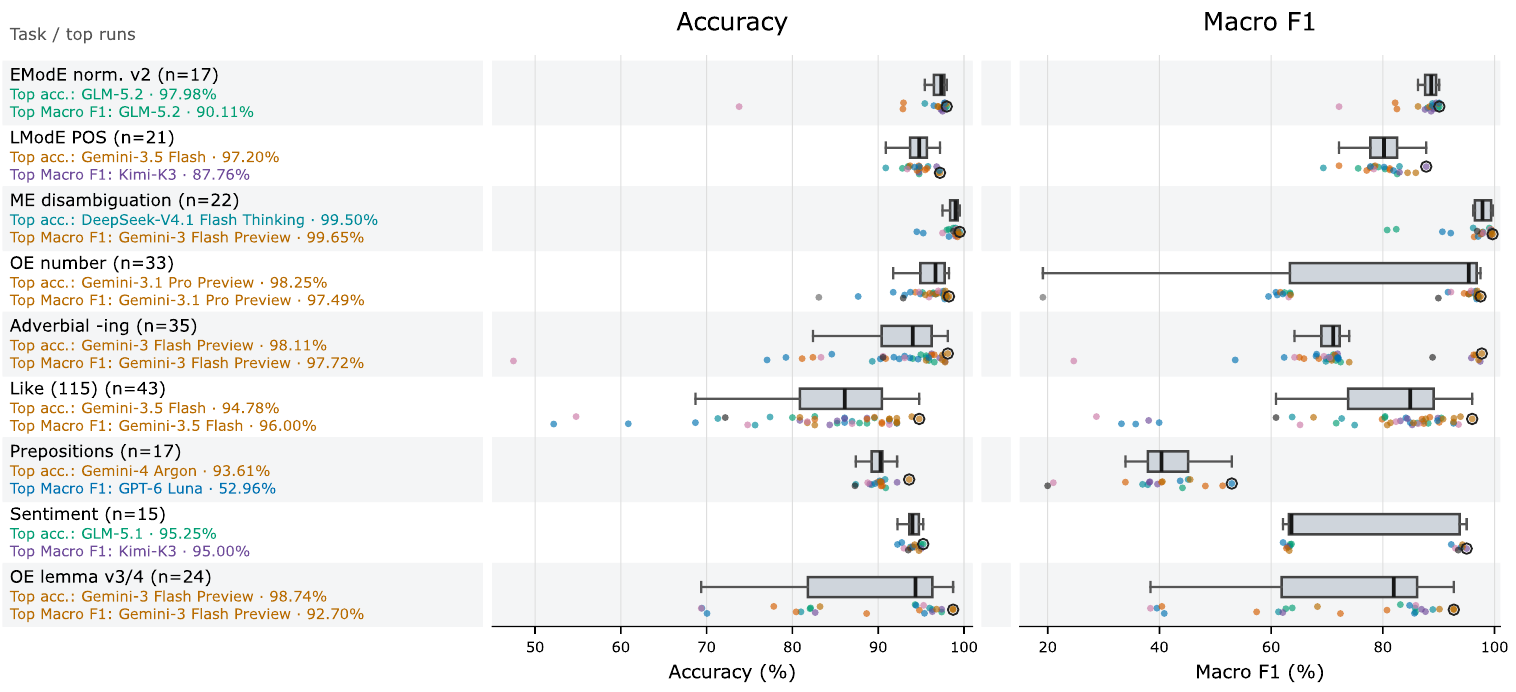}
\caption{{}Performance per task: Accuracy and Macro F1.}
\label{fig:06}
\end{figure}

Performance in individual tasks is shown in Figure~\ref{fig:06} listing all tasks, the number of runs, the best models and the individual runs as data-points colored by broad model families (grouped by their producers like Google or OpenAI) showing both overall Accuracy and Macro F1.

Some tasks, like ME disambiguation,\footnote{All models except gpt-oss-120 are over 95\% Accuracy and most are over 99\%.} have clearly reached saturation as benchmarks, or nearly so, like EModE normalization. These tasks may still be useful for benchmarking very small models but will probably need to be replaced in the future (individual tasks are discussed in section \ref{results-of-individual-tasks} in greater detail).

Top Accuracy or Macro F1 are not always the most important parameters --speed and especially efficiency may be paramount for large datasets. Figure~\ref{fig:07} shows the price-performance comparison, where estimated equal-task price is the average cost per prediction across the same nine tasks, with each task weighted equally regardless of dataset size or number of runs. Missing family--task prices were estimated using a two-way regression of log price on family and task effects, then converted back to mean prices with a lognormal correction; observed prices remained unchanged. This allows price and performance to be compared on the same task basis despite incomplete coverage. The dotted Pareto line joins families for which no other priced family is both cheaper and better, the green upper-left quadrant marks prices at or below the median and scores at or above the median among priced families. The whiskers follow the same methodology as in Figure~\ref{fig:05}.\footnote{For prices reported by the models' APIs, prices are included in their metric files and are valid as of the date of the test. Prices for models whose APIs did not return their prices, the pricing was harvested from public price lists of the providers in October 2026.}

\begin{figure}[htbp]
\centering
\includegraphics[width=1.0\linewidth,height=.78\textheight,keepaspectratio]{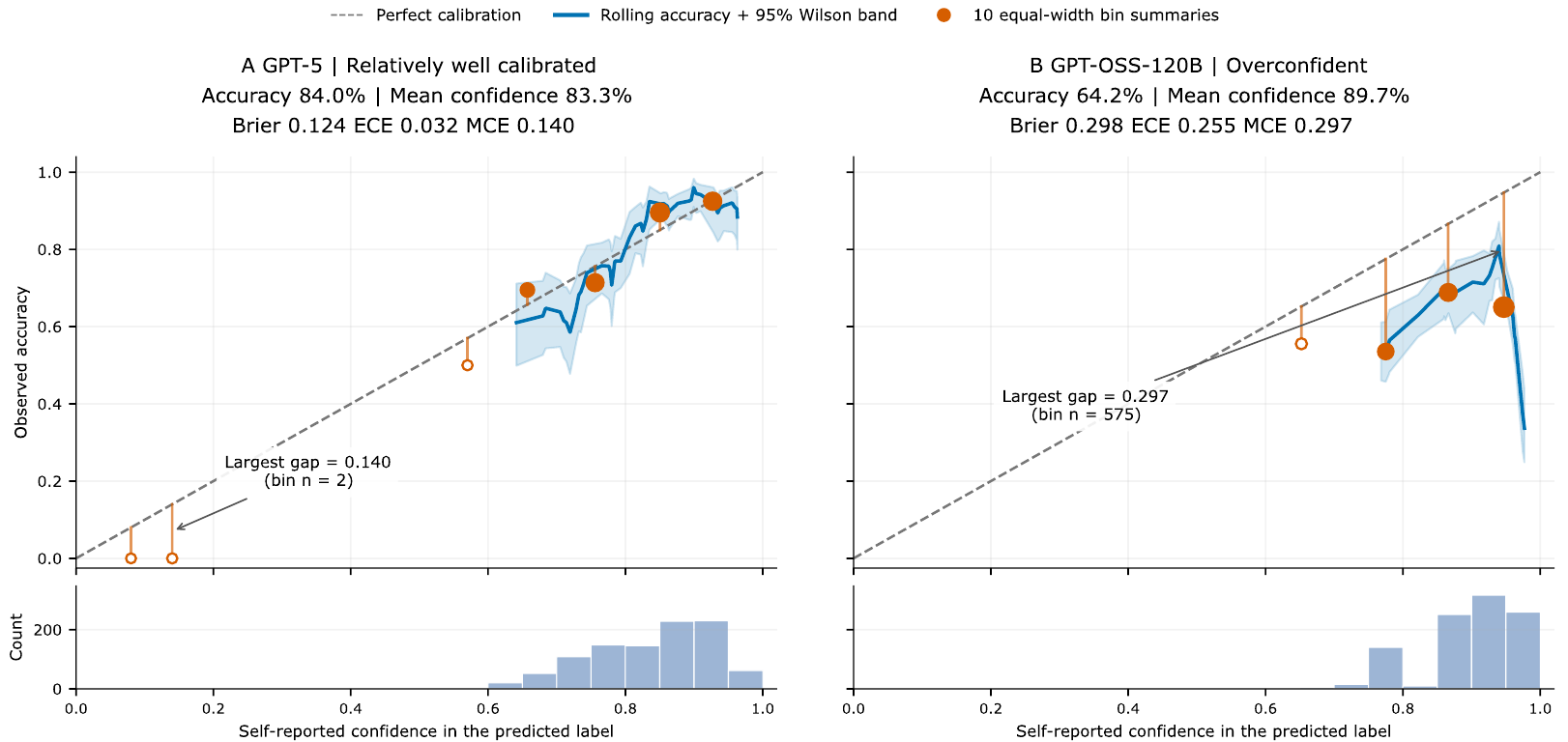}
\caption{{}Confidence calibration.}
\label{fig:08}
\end{figure}

However, when accuracy of the results is the most important parameter, it may be useful to identify items that were especially problematic for the annotator and perhaps manually correct them. To this end, \emph{dhAIBench} requests that models estimate their own confidence of each annotation {\citep{Miller2024-fa}.} This confidence is correlated with the actual observed accuracy, the maximum/expected confidence error (M/ECE) are calculated and confidence calibration charts are created for each run. Figure~\ref{fig:08} shows two confidence calibration charts with a well calibrated run on the left (where researchers can use model confidence to identify problematic items) and an overconfident run on the right (where the model is unable to identify problematic items).

\begin{figure}[htbp]
\centering
\includegraphics[width=.86\linewidth,height=.78\textheight,keepaspectratio]{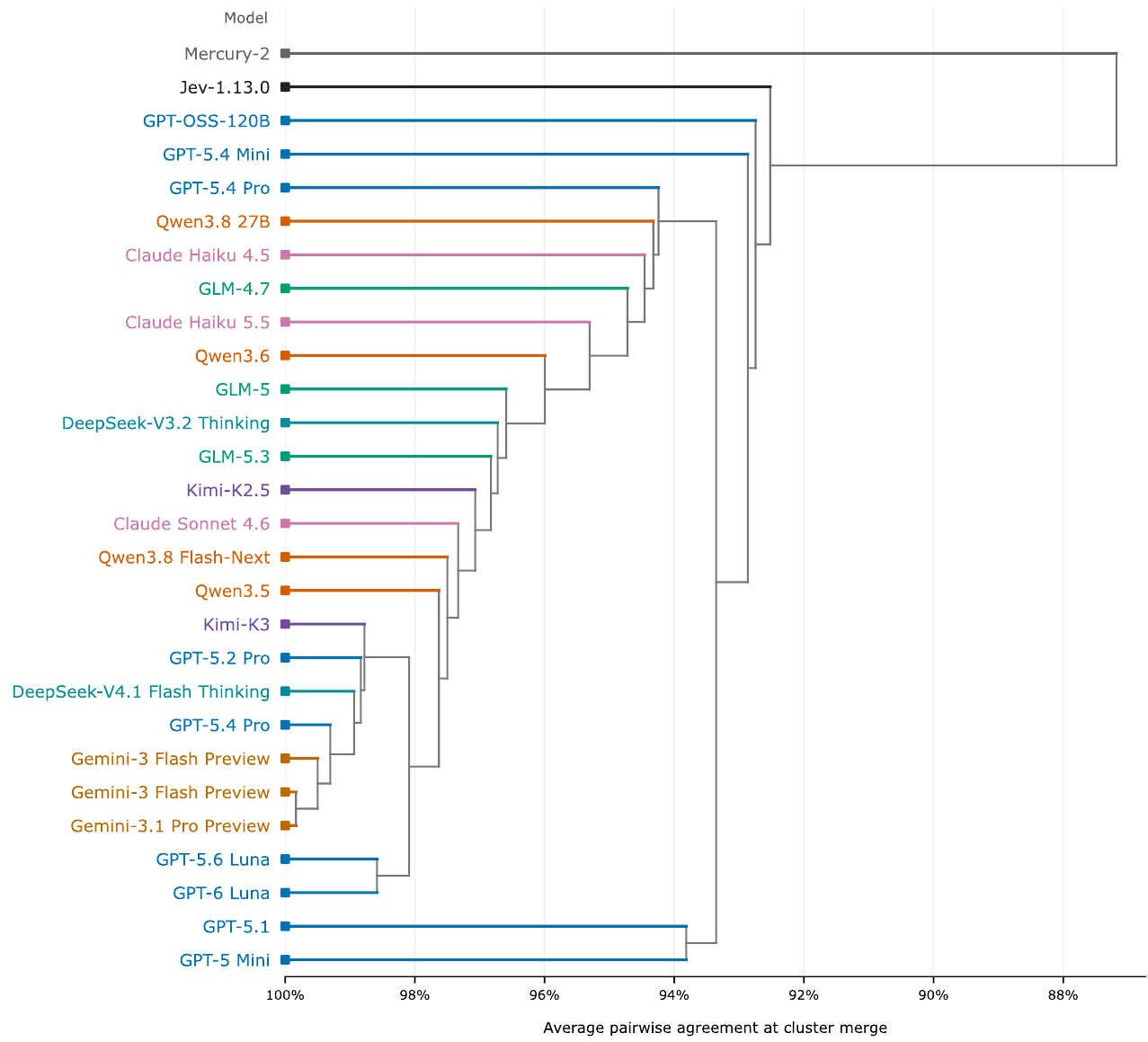}
\caption{{}Cross-model agreement for Old English number task.}
\label{fig:09}
\end{figure}

Our gold labels have so far been interpreted as the ground-truth, but they are in fact labels by fallible human annotators and Accuracy, should therefore be perhaps more accurately interpreted as inter-annotator agreement. This is calculated\footnote{Automatically after each run is completed but recalculation can be triggered manually, e.g. in case results are consolidated into one repository from multiple sources.} as cross-model Cohen's kappa, but users can add results by other human annotators as well. Figure~\ref{fig:09} shows how different models agree on the OE number tasks with model families mostly clustering near each other (highlighted in color).

\begin{figure}[htbp]
\centering
\includegraphics[width=1.0\linewidth,height=.78\textheight,keepaspectratio]{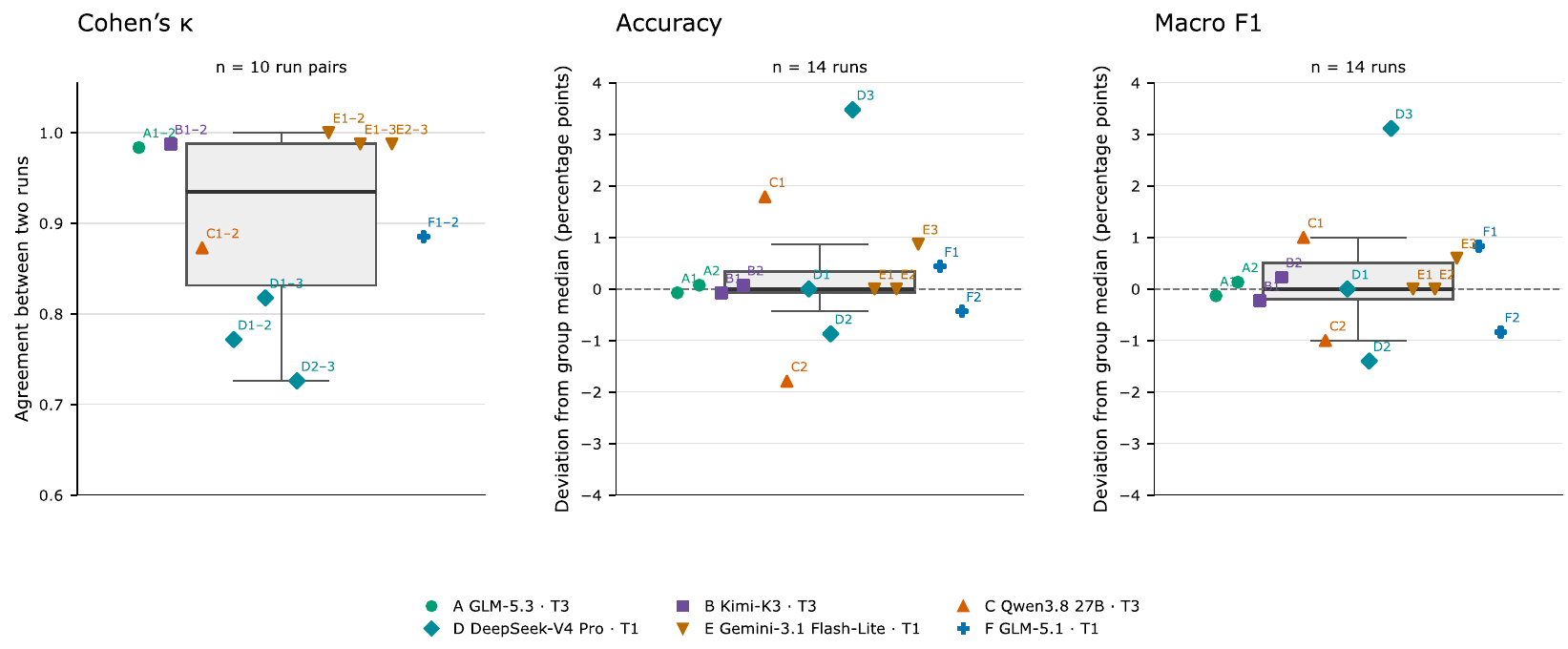}
\caption{{}Repeatability shown as differences in pairwise Cohen\textquotesingle s κ, Accuracy and Macro F1for the same tasks (T1 and T3) on the same models.}
\label{fig:10}
\end{figure}

When used with their default settings, LLMs are notoriously stochastic. The degree of repeatability and therefore to a degree reproducibility can be observed by analyzing the agreement between runs of the same model on the same task, calculated as pairwise Cohen's κ and by comparing Accuracy and Macro F1 of those runs. Gemini-3.1-Flash-lite and Kimi-k3 show better repeatability than GLM-5.1/3 or Qwen-3.8 27B (Figure~\ref{fig:10}).

\begin{figure}[htbp]
\centering
\includegraphics[width=1.0\linewidth,height=.78\textheight,keepaspectratio]{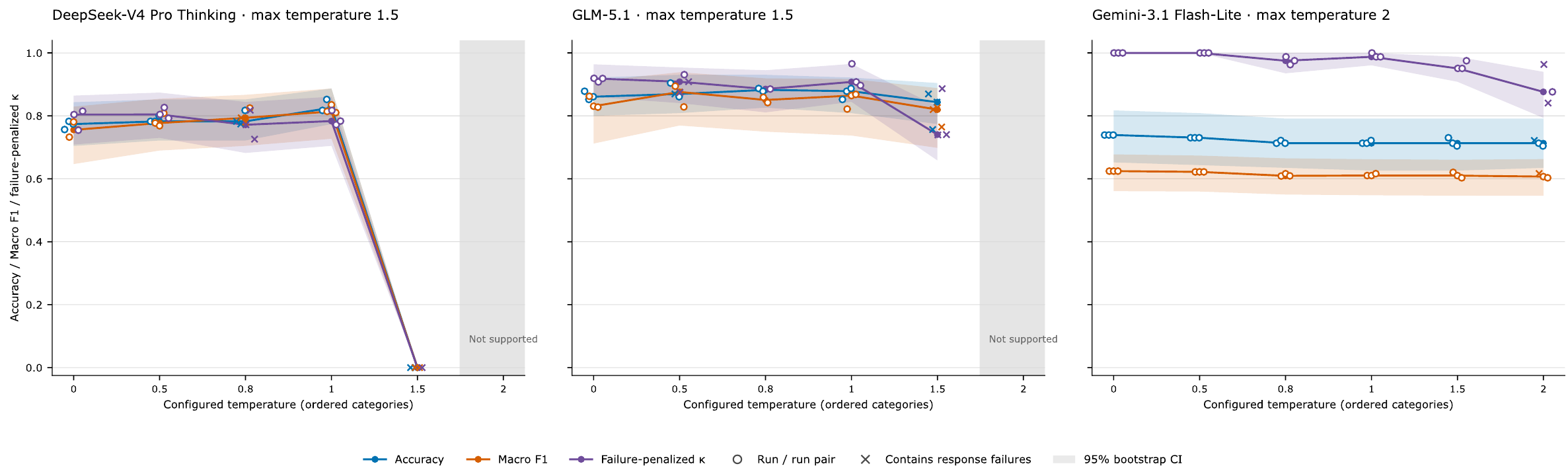}
\caption{{}The influence of temperature in T1 on Cohen's κ (agreement), Accuracy and Macro F1.}
\label{fig:11}
\end{figure}

One of the main model parameters influencing randomness of predicted tokens and therefore model stochasticity is temperature. We tested 6 temperature levels (0, 0.5, 0.8, 1, 1.5 and 2) on three models, see Figure~\ref{fig:11}. Across the three models, higher temperatures generally reduced repeatability, although accuracy and Macro F1 did not decline uniformly. DeepSeek-V4 Pro Thinking improved modestly from temperature 0 to 1, reaching median accuracy of 82.6\% and Macro F1 of 81.4\%, while κ remained around 0.77--0.80; at its maximum temperature of 1.5, however, both runs failed entirely, reducing all three measures to zero. GLM-5.1 maintained comparatively high accuracy (86--88\%), Macro F1 (83--88\%) and κ (0.89--0.92) through temperature 1, but at 1.5 its κ fell to 0.74, quality declined modestly, and 28 of 345 responses failed. Gemini-3.1 Flash-Lite was perfectly repeatable at temperatures 0 and 0.5 (κ = 1), despite lower prediction quality; accuracy and Macro F1 declined slightly and then remained broadly stable, while κ fell to 0.88 at temperature 2, with only one failed response. The failed runs were largely due to the models' inability to form valid JSONs at higher temperatures. The provider defaults, 0.8 for DeepSeek and GLM, and 1 for Gemini, therefore, lie within relatively stable operating ranges, but were not uniformly the best-performing settings in these runs.

\begin{figure}[htbp]
\centering
\includegraphics[width=1.0\linewidth,height=.78\textheight,keepaspectratio]{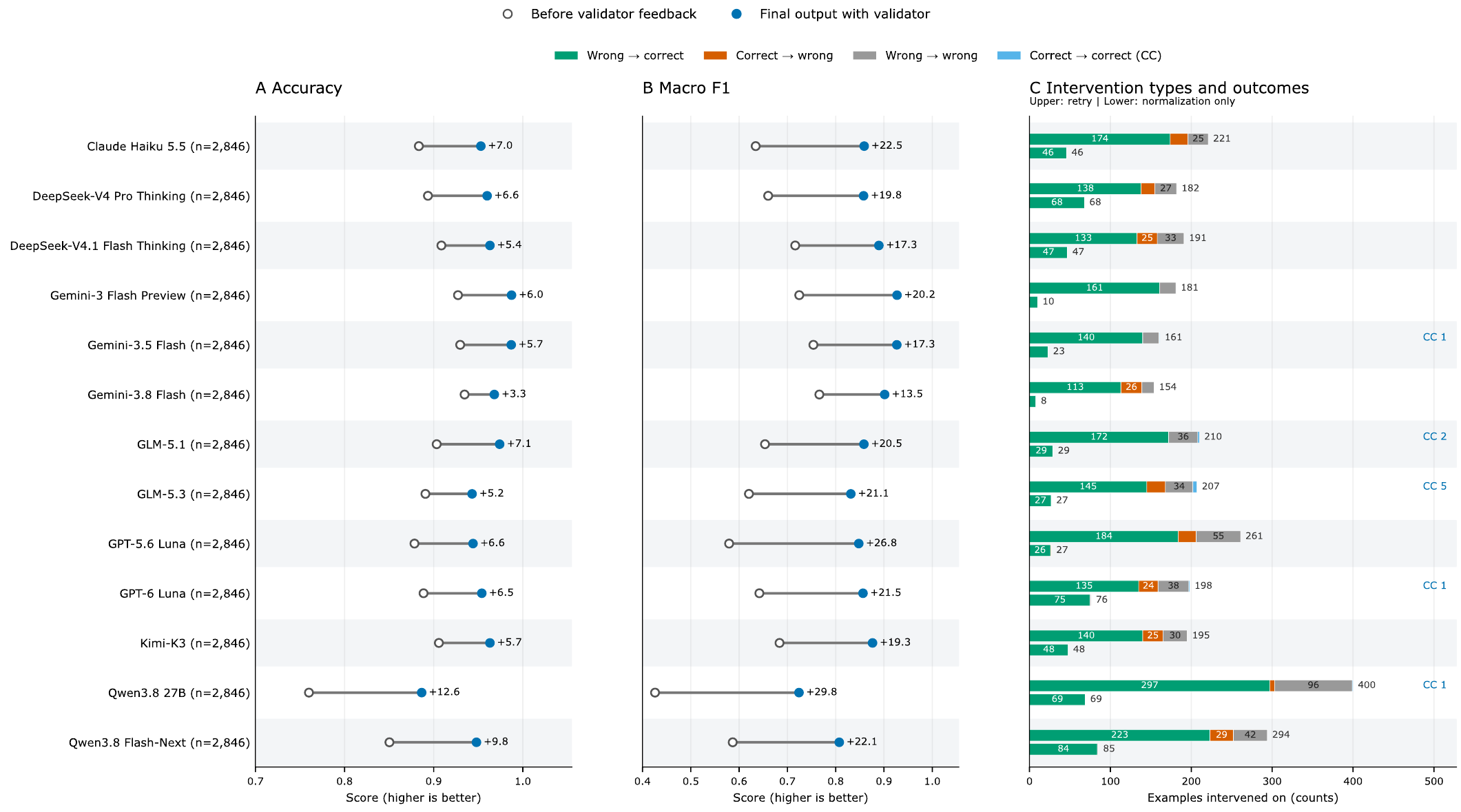}
\caption{{}Validator in T7: quality gains in Accuracy and Macro F1 and intervention types.}
\label{fig:12}
\end{figure}

A particularly common worry when using LLMs for annotations are hallucinations. These are relatively easy to avoid with a small tagset, but in case of a large pool of permissible annotation, \emph{dhAIBench} delegates the task to external validators (see section \ref{validators}). We have tested the validator on the OE lemmatization task, given the tag set comprises of all OE dictionary headwords, and we compared each run's first pre-validator prediction with its final output. Figure~\ref{fig:12} shows that validator feedback improved accuracy and Macro F1 in all 13 included Old English lemmatization runs, with median gains of 6.54 and 20.47 percentage points, respectively. Qwen3.8 27B showed the largest improvements (+12.65 points in accuracy and +29.81 in Macro F1), although its final scores remained below those of the other models; Gemini-3 Flash Preview achieved the highest final accuracy (98.74\%) and Macro F1 (92.70\%). Across 36,998 predictions, 3,418 received an intervention: 2,855 involved retries and 563 normalizations alone. Interventions corrected 2,715 initially wrong predictions but changed 220 correct predictions to incorrect ones; normalization alone was particularly successful, correcting 560 cases and harming only one. Macro F1 gains were consistently larger than accuracy gains, partly because correcting invented lemmas also removed spurious predicted classes.

\subsection{Results of individual tasks}\label{results-of-individual-tasks}

The metrics in \ref{general-results} show how well the models perform, but not where and why they fail. This section therefore offers a qualitative and comparative look at selected tasks, based on an inspection of the mismatches between model predictions and the reference annotations. The aim is not a systematic error analysis but to complement the quantitative results and give the reader a sense of where the inaccuracies typically arise.

\subsubsection{\texorpdfstring{T1: Discourse-pragmatic uses of \emph{like}}{T1: Discourse-pragmatic uses of like}}\label{t1-discourse-pragmatic-uses-of-like-1}

The top results for this task were around 95\% for both Accuracy and Macro F1. This is lower than the 98\% reported by Raušová (by private communication) for her expert human annotators, but higher than the 84.5\% Accuracy reported by Popescu-Belis \& Zufferey {\citeyearpar{Popescu-Belis2011-mp},} who used decision trees to automate the binary annotation discourse-marker/other. In our case, the problems mostly concentrated in misidentification of Discourse Markers as clause-medial or clause-final markers or with misidentification of clause-final as clause-medial markers. This is similar across all models including those with significantly worse results overall. This seems to be largely due to irregular punctuation and the disconnected nature of the transcript where speakers talk over each other. This makes it difficult for the model to identify clause boundaries and consequently choose the wrong annotation but could possibly be alleviated by alleviated by larger context and better instructions for parsing this type of transcript.

\subsubsection{\texorpdfstring{T2: Adverbial \emph{-ing} clauses}{T2: Adverbial -ing clauses}}\label{t2-adverbial--ing-clauses-1}

\begin{figure}[htbp]
\centering
\includegraphics[width=.80\linewidth,height=.78\textheight,keepaspectratio]{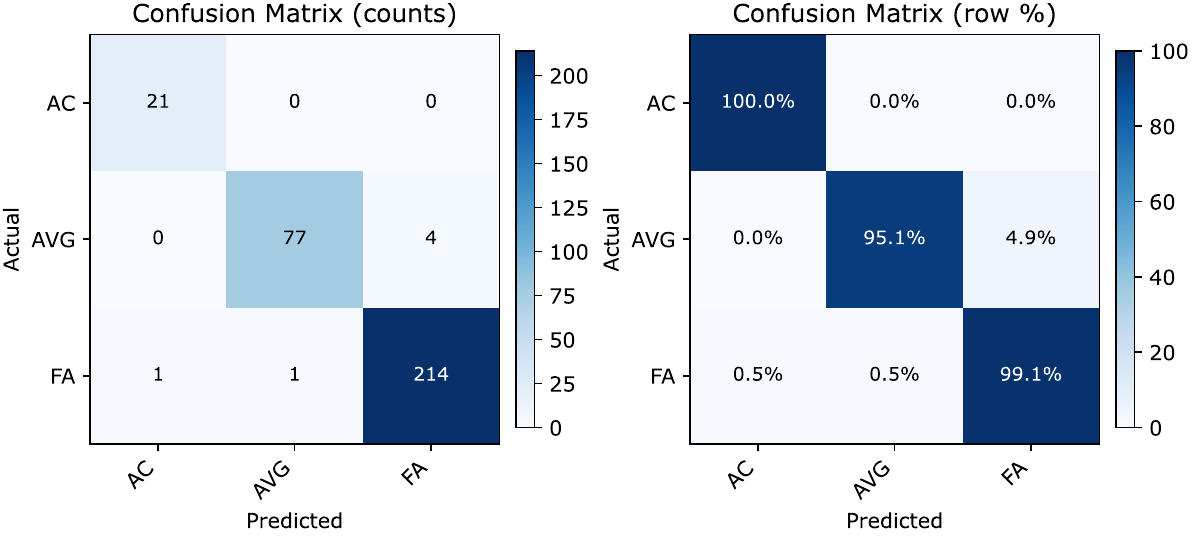}
\caption{{}The confusion matrix for the top run of T2 using gemini-3-flash-preview in counts and percentages.}
\label{fig:13}
\end{figure}

The best results for this task reached around 98\% in Accuracy and Macro F1, i.e. only 6 misclassifications out of 318 items for the best run (see the confusion matrix in Figure~\ref{fig:13}, which is automatically created for every run).

Since even small models like qwen3.8-27b scored above 90\%, the test is largely saturated. The errors were mostly adverbial verbal gerunds (AVG) classified as free adjuncts (FA) or unclassified, and this was common for all the models. While we do not have comparable data for human or automatic annotators, we have discovered 2 misclassifications and 15 potentially ambiguous instances among our gold labels, i.e. the top run is comparable to an expert human annotator.

\subsubsection{T3: Normalization of Early Modern English printed text}\label{t3-normalization-of-early-modern-english-printed-text-1}

Accuracy across the models on the EModE normalization task was generally above 97\%, with glm-5.2 achieving the best performance (97.98\%) and gpt-oss-120b:free the worst (32.35\%). The poor performance of gpt-oss-120b:free was due to the format of its output. The prediction column contained classification labels, such as \emph{normalization}, \emph{no\_change}, and \emph{unclassified}, rather than the actual normalized word forms. In some cases, the expected forms appeared only in the model\textquotesingle s explanations, not in the structured output. This made it impossible to directly evaluate the model\textquotesingle s spelling normalization accuracy automatically.

An analysis of the mismatches in the best-performing run revealed that, despite its generally high accuracy, the model did not consistently follow certain instructions in the prompt. One recurring issue was the failure to follow Rule 7, which states that proper names should not be normalized or changed. For example, the model changed \emph{Iuppyter} to \emph{Jupiter}. Although the resulting spelling is correct in modern English, the change violates the explicit instruction to preserve proper names. Problems also occurred with Latin expressions, apostrophes, and corrupted characters. Conversely, one example of a prediction that was marked as incorrect but may actually be acceptable is \emph{oft}, which was standardized to \emph{often} in the truth column. However, \emph{oft} is also a legitimate English variant, has its own entry in the OED, and is closer to the original form. Another ambiguous case involves words that can be normalized in more than one way. For example, \emph{Gayle} could potentially be normalized to either \emph{gaol} or \emph{jail}. Both forms are valid, although each is supported by a slightly different rationale.

When Hämäläinen et al. {\citeyearpar{Hamalainen2018-li}} compared commonly used digital tools for text standardization with several alternative methods, including NMT, SMT, edit distance, and the rule-based finite-state transducer (VARD2), they concluded that NMT appeared to be the most effective individual method, achieving an accuracy of 46\%. However, this result still falls considerably short of the performance achieved using LLMs.

Similarly, texts from the Early Print Database, standardized using MorphAdorner software in combination with manual editing, were found to have a higher relative error rate (5.26) than those standardized using LLMs (0.62--1.87) {\citep{Titlbachova2025-fr}.}

\subsubsection{T4: Error-correction of prepositions in learners' language}\label{t4-error-correction-of-prepositions-in-learners-language-1}

Accuracy across the models on the L2 prepositional use classification task ranged from 87.3\% (jev-1.13.0) to 93.61\% (gemini-4-argon). An analysis of the mismatches in the most successful run showed that, despite generally high accuracy, the model appeared to apply more inclusive criteria of acceptability than the human annotator. A particularly frequent source of disagreement involved prepositional uses that, although restricted to certain contexts, are attested and were therefore retained by the model as grammatically acceptable (e.g. `I have been in Belfast'; `buy something in McDonald's'). However, the model also left some questionable cases uncorrected (e.g. `spend so much time in reading a novel'; `go in a pub'), while some of the annotator's corrections reflected only differences between regional varieties of English or purely stylistic preferences (e.g. `a tour in the centre of Europe' corrected to `a tour around the centre of Europe'). Only rarely did the disagreements result from an erroneous identification by either the annotator or the model. Rather, most of these discrepancies appear to stem from differences in the scope of acceptability applied by the model and the annotator as well as from the lack of detailed inclusion criteria in the original annotation. The remaining cases were peripheral and involved instances in which the model and the annotator proposed different but equally acceptable corrections or interpreted fragmented speech differently.

\subsubsection{T5: Part-of-speech tagging of Late Modern English correspondence}\label{t5-part-of-speech-tagging-of-late-modern-english-correspondence-1}

While the top Accuracy achieved for this task is around 97\%, Macro F1 is only around 87\%, which is primarily caused by a large number of relatively rare tags. This compares well with the previously reported 94.5\% Accuracy {\citep{Saario2021-mm}} for the \emph{Tagged Corpus of Early English Correspondence Extension} achieved by first normalizing the corpus and then tagging it with CLAWS. The present dataset was shared by Saario-Ramsay et al. {\citep{Saario-Ramsay_undated-jh}} who achieved 94.28\% using CLAWS and 92.26\% using custom tuned smaller LLMs noting that ``the best-performing model is the largest one, but the difference between 72B and 7B models is only 0.7pp''. In our results, the difference between the top scoring gemini-3.5-flash and the smallest tested model qwen3.8-27b is 2.46\%. Similar to Saario-Ramsay et al., the worst results were generally achieved for the letter written by a lower-class woman (PAUPER\_11)\footnote{96.3\% for the top run compared to 84.16\% for the top LLM and 77.38\% for CLAWS as reported by Saario-Ramsay et al. But note that for the best runs, the letters BLOMEFI and PINNEY actually achieved marginally worse results.} and the difference between the best and worst per letter results was much smaller than the difference they reported for CLAWS indicating that the LLMs are particularly more capable compared to more traditional software like CLAWS when dealing with highly non-standard texts.

Since we had both the training and test subsets of letters available from Saario-Ramsay et al., we have also experimented with various prompts to achieve both good Accuracy, but especially to improve the Macro F1 values. We have included the complete CLAWS7 guide including their examples of ambiguous or rare tag usage and we have also included the complete training subset. The Accuracy was roughly comparable, but Macro F1 was about 10\% better when a training dataset was submitted in the prompt than with the detailed CLAWS7 guidelines.

\subsubsection{T6: Semantic disambiguation of Middle English nouns}\label{t6-semantic-disambiguation-of-middle-english-nouns-1}

All models except for gpt-oss-120b reached over 95\% Accuracy and the top runs reached over 99\% in both Accuracy and Macro F1showing that this test is saturated for the current generation of LLMs. We can compare the results to our two human annotators who reached Accuracy of 93\% (experienced postgraduate student of Middle English) and 99\% (professor in the field of ME) showing the models are now on par with the top human annotators.

\subsubsection{T7: Lemmatization of Old English}\label{t7-lemmatization-of-old-english-1}

Accuracy on the Old English lemmatization task ranged from 69.37\% to 98.74\% across the recorded runs. Gemini-3 Flash Preview achieved the highest accuracy, agreeing with 2,810 of the 2,846 reference annotations, and the highest Macro F1 (92.70\%). Differences between initial predictions and outputs following lexical validation are discussed with Figure~\ref{fig:12}.

Qualitative assessment nevertheless revealed several difficulties arising from the relationship between the corpus annotations and the lexicon used by the validator. A number of headwords used as gold labels are absent from the validator lexicon, including \emph{mann}, \emph{menniscnes} and \emph{gescyldan}. Other lemmas occur in the lexicon a different POS category: \emph{swilce}, for example, is listed as an adverb but also functions as a preposition in the corpus, while \emph{se} lacks a determiner entry. The lexicon also retains alternative headword spellings, including \emph{self/sylf}, \emph{ful/full} and \emph{andgit/andgyt}. Consequently, instead of standardizing the predictions, the validator may confirm a variant prediction causing a mismatch against the gold labels. A further complication is the use of some slash-separated gold labels, such as \emph{heofon/heofone} and \emph{wille/willa}: which cannot match with single lemma predictions.

Some disagreements also expose inconsistencies or apparent errors in the gold labels. Latin \emph{pater} is tagged both FW and NR, while \emph{noster/nostre} likewise receive different treatment, affecting the applicability of the instruction to return \emph{foreign\_word} for FW tokens. \emph{Marian}, referring to Mary, receives the reference label \emph{name}, rather than a lexical headword or the designated foreign-word label. Other apparent errors of the gold labels used include \emph{geþeoþ} assigned to \emph{geþeodan} rather than \emph{geþeon}, \emph{elles} `otherwise' assigned to \emph{eall}, and three instances of \emph{fet} `feet' assigned to \emph{fæt} `vessel' rather than \emph{fot}. These cases reinforce the interpretation of accuracy as agreement with the supplied annotations rather than an independent measure of linguistic correctness.

Tichý's {\citeyearpar{Tichy2017-ax}} rule-based morphological analyzer achieved 95\% recall, rising to 97.3\% with spelling-variation filters, but these figures measure inclusion of the correct analysis among potentially multiple alternatives rather than selection of a single contextually appropriate lemma. Our top Accuracy is substantially higher than the approximately 72\% reported by Kawakami and Kawanami-Hanpeter {\citeyearpar{Kawakami2022-cr}} for a rule-based lemmatizer using context-sensitive weighted edit distance on Ælfric's \emph{De Sancta Eugenia}. They also exceed the 67.58\% held-out test accuracy reported in the documentation accompanying Arista's {\citeyearpar{Arista2026-gk}} character-level Transformer. These are contextual comparisons rather than equivalent evaluations: the studies differ in their samples, reference conventions and available linguistic information.

\subsubsection{T8: Number of Old English nouns}\label{t8-number-of-old-english-nouns-1}

The top result for this task is around 98\% for both Accuracy and Macro F1, but smaller models perform significantly worse, c.f. qwen3.8-27b with 61\% for both metrics, suggesting this test is still useful for less powerful models. Most of the differences between the predictions and the gold labels in the top runs involved the noun \emph{hand} and specifically the OE construction \emph{to handa} expressing possession rather than physical hand(s) and therefore semantically difficult to interpret in terms of number. The models` explanations reflect this and note that based on the formal interpretation, this is more probably dative singular than accusative plural. This is philologically sound explanation and should we accept it, the results reach over 99\%.

In comparison, Domínguez Barragán {\citeyearpar[297]{Dominguez_Barragan2024-jp}} reports approximately 90\% performance for automatic assignment of grammatical number using a pretrained spaCy pipeline. This provides a broader morphological comparison, since the evaluation is not restricted to number-ambiguous nouns.

\subsubsection{T9: Sentiment analysis of the Movie Review Data}\label{t9-sentiment-analysis-of-the-movie-review-data-1}

The overall success rate of the models' predictions agreeing with the human annotators was ranging from 92.25\% (gpt-5.4-mini) to 95.25\% (glm-5.1). Inspection of the cases where the predictions did not agree showed that most of these statements contained both negative and positive aspects, so the decision was rather subjective. For example, the statement ``light, cute and forgettable'' was assessed by the human annotator as positive but by glm-5.1 as negative, with the model's explanation: ``While `light' and `cute' provide mild positive qualifiers, the conclusive term `forgettable' acts as a strong negative judgment on the movie's lasting value, outweighing the slight praise and resulting in an overall negative sentiment.'' gpt-5.4-mini also assessed it as negative, with a similar explanation: ``The node describes the movie as `forgettable,' which is a negative. Although `light' and `cute' are mildly positive, the overall assessment is dismissive and unfavorable.'' This suggests that the errors come mainly from items with mixed sentiment, not from an underspecified prompt or a failure of the models.

\FloatBarrier

\section{Conclusions}\label{conclusions}

Our results indicate that LLMs represent a competitive tool for various NLP tasks -- both in tasks where other NLP tools have been successful as well as in those that have so far resisted automatic annotation. We have shown competitive results for traditional tasks such as PoS annotation or lemmatization, specifically for highly varied and low-resource varieties (e.g. historical varieties of English) and for tasks highly reliant on semantic and contextual analysis, such as discourse-pragmatic categorization or semantic disambiguation. This does not mean that LLM annotation is suitable for every linguistic task, requiring word-internal segmentation, where difficulties in sub-token processing {\citep{Hiraoka2025-du}} may limit performance (see section \ref{future-research}).

The efficiency of using LLMs for annotation has to be, however, assessed individually for each tasks. While it is difficult to estimate how resource intensive especially the commercial frontier models are,\footnote{Their authors publish little information on energy or water consumption, but price can be a useful proxy: more expensive models are probably more resource intensive.} it is clear that they are in general more resource intensive than traditional deterministic NLP tools. Models we have benchmarked so far show large differences in prices: from Jev-1.13.0 at \$0.00002 to gpt-5.4-pro at \$0.28 estimated price per prediction.\footnote{These are different from advertised price per input/output tokens and are estimated from our own results and based on the total expense per model divided by its total predictions (annotation instances). They are influenced by the verbosity of the particular model, its ``effort'' or ``thinking levels'', pricing tier, token caching etc.} In other words, to lemmatize a corpus of 100 mil. Tokens can cost as little as \$500 with Jev\footnote{Note that while we ran all open-weights models for free on a shared research infrastructure (e-INFRA CZ) and smaller models like qwen3.8-27b can easily run on current consumer hardware, this does not mean these do not consume resources and we have therefore used their listed prices with other providers used in this study to estimate their prices.} or as much as \$28 mil. with gpt-5.4-pro.

Therefore, LLMs should be used for annotation only when necessary and the less resource intensive (cheaper) models used in cases where their results are sufficient for the task at hand. However, it seems that in all our tasks, manual annotation would have been less efficient and only marginally more accurate in some of the tasks. Therefore, for our tasks, we can answer the question in Kuzman et al.'s paper \emph{ChatGPT: Beginning of an End of Manual Linguistic Data Annotation?} {\citeyearpar{Kuzman2023-es}} with affirmative.

\FloatBarrier

\section{Future research}\label{future-research}

As noted above, a number of authors (Saario-Ramsay et al., Faulkner \& Magnanti) have already joined the benchmarking effort and some are currently using dhAIBench for their research (Smažíková \& Kratochvílová are using the tool to semantically annotate Spanish modal periphrases). This should extend the benchmarks to other languages.

To test annotations requiring sub-token segmentation, we plan benchmarking annotations of Old and Middle English alliteration and multilingual morphological segmentations.

We also plan to extend the tool to handle more complex inputs and outputs, particularly XML, to facilitate features such as automating annotations according to Text Encoding Initiative (TEI) schemas.

Finally, we invite others to join: to look at our website and assess whether there are models suitable for their task, run the benchmarks themselves or approach us to run it for them, and use the tool to annotate their datasets.

\FloatBarrier
\section*{References}
\phantomsection
\addcontentsline{toc}{section}{References}

This work was supported by the Ministry of Education, Youth and Sports of the Czech Republic through the e-INFRA CZ (ID:90254).


\end{document}